\documentclass[pmlr]{jmlr}

 \usepackage{booktabs}
\usepackage{longtable}
\usepackage[load-configurations=version-1]{siunitx} 

\makeatletter
\def\set@curr@file#1{\def\@curr@file{#1}} 
\makeatother

\makeatletter
 \let\Ginclude@graphics\@org@Ginclude@graphics 
\makeatother

\theorembodyfont{\upshape}
\theoremheaderfont{\scshape}
\theorempostheader{:}
\theoremsep{\newline}

\jmlrvolume{[VOLUME \# TBD]}
\jmlryear{2024}

\title[LLM based OPQRST Extraction]{Extracting OPQRST in Electronic Health Records using Large Language Models with Reasoning}

\author{\Name{Zhimeng Luo}
       \Email{zhl123e@pitt.edu}\\ 
       \addr School of Computing and Information\\
       University of Pittsburgh\\
       Pittsburgh, PA, USA 
       \AND
       \Name{Abhibha Gupta}
       \Email{abg96@pitt.edu}\\ 
       \addr School of Computing and Information\\
       University of Pittsburgh\\
       Pittsburgh, PA, USA
       \AND
       \Name{Adam Frisch}
       \Email{frischan@upmc.edu}\\ 
       \addr Department of Emergency Medicine\\
       University of Pittsburgh\\
       Pittsburgh, PA, USA
       \AND
       \Name{Daqing He}
       \Email{dah44@pitt.edu}\\ 
       \addr School of Computing and Information\\
       University of Pittsburgh\\
       Pittsburgh, PA, USA
       }

\begin{document}

\maketitle

\begin{abstract}
The extraction of critical patient information from Electronic Health Records (EHRs) poses significant challenges due to the complexity and unstructured nature of the data. Traditional machine learning approaches often fail to capture pertinent details efficiently, making it difficult for clinicians to utilize these tools effectively in patient care. This paper introduces a novel approach to extracting the OPQRST assessment from EHRs by leveraging the capabilities of Large Language Models (LLMs). We propose to reframe the task from sequence labeling to text generation, enabling the models to provide reasoning steps that mimic a physician's cognitive processes. This approach enhances interpretability and adapts to the limited availability of labeled data in healthcare settings. Furthermore, we address the challenge of evaluating the accuracy of machine-generated text in clinical contexts by proposing a modification to traditional Named Entity Recognition (NER) metrics. This includes the integration of semantic similarity measures, such as the BERT Score, to assess the alignment between generated text and the clinical intent of the original records. Our contributions demonstrate a significant advancement in the use of AI in healthcare, offering a scalable solution that improves the accuracy and usability of information extraction from EHRs, thereby aiding clinicians in making more informed decisions and enhancing patient care outcomes.
\end{abstract}

\section{Introduction}

The accurate and efficient extraction of critical patient information from Electronic Health Records (EHRs) is an important problem in machine learning and healthcare. In healthcare, the ability to swiftly and accurately extract pertinent information from EHRs is crucial, yet the complexity and volume of these records make this task difficult. Clinicians rely on these details to make informed decisions about patient care, and the ability to quickly identify key elements such as the OPQRST (a common mnemonic for patient assessment, which contains Onset, Provocation/palliation, Quality, Region/Radiation, Severity, and Time) can greatly improve healthcare delivery. However, addressing this problem is challenging due to the vast amount of complex and unstructured data present in EHR notes, making it difficult for traditional machine learning approaches to effectively capture and extract the relevant information.

Previous attempts~\citep{10313411} to tackle the extraction of OPQRST from EHRs have been treated as a sequence labeling task, and limited by the lack of labeled training data in this low-resource setting, hindering the performance of pre-trained models. Additionally, these models often function as black boxes, lacking the ability to provide reasoning for their outputs, which is crucial in high-risk medical scenarios. In contrast, we propose a novel approach to extract OPQRST in EHRs by leveraging Large Language Models (LLMs) such as Llama 2~\citep{touvron2023llama} and GPT-4~\citep{achiam2023gpt}. By framing the task as a text generation problem instead of sequence labeling, we enable the model to provide reasoning steps, mimicking the physician's thought process. This method not only aids in mimicking a physician's cognitive process during patient evaluation but also enables the model to learn with minimal labeled data, focusing on reasoning rather than only extraction. This approach not only improves interpretability but also reduces the need for extensive ground truth annotations, as the model's output can be guided by analyzing the generated reasoning.

Furthermore, considering the unique challenges posed by viewing the task as text generation rather than sequence labeling, the problem with the generative model is that it tends to output text that is similar but not exactly the same as the original text, whereas the traditional Named Entity Recognition (NER) evaluation metrics only designed to capture the hard matches, resulting in the categorization of any phrases not directly extracted from the original text as incorrect. To tackle this problem, we introduce a novel adaptation to the traditional NER evaluation metrics by adding semantic similarity checking for the generated phrases. We employ metrics like the BERTScore~\citep{ZhangKWWA20} or prompt LLMs to assess the semantic similarity between the generated phrase and ground truth phrase. This approach allows us to capture the effectiveness of the generated outputs in reflecting the true clinical context, thereby bridging the gap between machine-generated phrases and their human-generated counterparts.

Our contributions are as follows:
\begin{itemize}
\item Addressing the important problem of extracting OPQRST in EHRs, which is crucial for clinicians to make informed decisions about patient care.
\item Framing the task as a text generation problem instead of sequence labeling, enables the leverage of Large Language Models (LLMs) to tackle the task in a few-shot setting and provide reasoning steps to mimic the physician's thought process, thereby improving interpretability and reducing the need for extensive ground truth annotations.
\item Extending the traditional Named Entity Recognition (NER) evaluation metrics, i.e., entity-based F1 score, by proposing a new metric that measures the similarity between the generated phrase and the ground truth using BERTScore or PromptLLM, allowing for a more comprehensive assessment of the model's performance in cases where the generated text is not an exact match to the original.
\end{itemize}

\subsection*{Generalizable Insights about Machine Learning in the Context of Healthcare}


\begin{itemize}
\item Reframing Problem Definition: Transitioning the task of extracting critical patient information from sequence labeling to text generation represents a significant shift in approach. This redefinition enables the use of advanced Large Language Models (LLMs), which are capable of generating reasoning steps akin to a physician’s thought process. This insight emphasizes the importance of aligning machine learning tasks with the cognitive workflows of clinicians, enhancing both the usability and acceptance of AI tools in clinical settings.

\item Modeling for interpretability and low-resource efficiency: By leveraging LLMs in a few-shot learning framework, this study illustrates how machine learning can be effectively applied even with minimal labeled datasets. This approach addresses the common challenge in healthcare NLP of scarce or expensive annotations.

\item Adaptation of Evaluation Metrics: The introduction of a novel adaptation to traditional NER metrics, incorporating semantic similarity assessments (e.g., BERTScore, PromptLLM), showcases a method to better evaluate AI performance in clinical contexts. This development is significant as it addresses the limitations of existing metrics by recognizing the contextual accuracy of machine-generated text, even when exact text matches are absent.
\end{itemize}

\section{Related Work}


Named Entity Recognition (NER) is a common NLP technique for locating entity mentions in a text. Many review studies~\citep{li2023far,landolsi2023information, hahn2020medical, WANG201834} have been conducted to examine the entities in medical text and develop automatic methods to extract them.

\subsection{Medical Information Extraction}
The field of medical information extraction (IE) is crucial for distilling structured insights from the vast expanses of unstructured EHRs. Recent literature review~\citep{WANG201834} highlights the critical need for collaboration between natural language processing (NLP) experts and clinical professionals to address the uniqueness of clinical data. This review underscores the scarcity of publicly available clinical datasets—a consequence of stringent privacy regulations like the Health Insurance Portability and Accountability Act (HIPAA) and institutional privacy concerns. Concurrently, \cite{landolsi2023information} emphasizes the essential tasks within clinical IE, such as NER and relation extraction (RE). These tasks are fundamental for transforming narrative clinical texts into actionable data, supporting medical decision-making and risk prediction. This work also points to the inherent challenges posed by the linguistic complexities of clinical narratives, including negation, co-references, and the variability of medical terminology, which significantly complicate the IE process. \citet{zhao2019neural} developed a multi-task framework to jointly tackle the task of medical NER and entity normalization. \citet{bhatia2019comprehend} developed a web service for medical NER and relationship extraction in medical data. To tackle the issue of limited data annotation, \citet{hofer2018few} proposed a method for few-shot learning of NER in the medical domain.

\subsection{LLMs based Extraction}
In-context learning~\citep{min2022rethinking}, which relies on prompt engineering to guide large pre-trained language models, has shown promising results in clinical NLP tasks where labeled datasets are often limited~\citep{agrawal2022large}. In clinical NLP, prompt engineering is crucial for leveraging state-of-the-art language models, as labeled datasets are often scarce, expensive, time-consuming to create, and constrained by data use agreements across institutions. \cite{wang2023large} highlight the need for addressing specific challenges such as wording ambiguity, lack of context, and negation handling while emphasizing the importance of responsible LLMs implementation and collaboration with domain experts in healthcare settings. Different types of prompts proposed in recent literature, including prefix ~\citep{li2021prefix}, cloze~\citep{liu2023pre}, chain of thought~\citep{NEURIPS2022_9d560961}, and anticipatory prompts~\citep{hancock2019learning}, have been used under the clinical settings~\citep{info:doi/10.2196/55318}.

\section{Methods}

\subsection{Prompt Creation}

The designing of the extraction prompt involved several steps as there were many parts to it. We engineered it in an iterative manner. We had prior knowledge about the dependency between the OPQRS entities and the chief complaint. A chief complaint (CC) is an initial statement of patient-derived medical issues, which is often elicited prior to formal medical tests and diagnoses. It provides a brief statement about a patient's reasons for encounter, current symptoms, and medical history~\citep{chang2020generating}. Following queue from the paper {cite question decomposition paper}  and our prior knowledge we initially crafted an initial prompt that would decompose the entity extraction task into question-answer (Q/A) pairs. The first Q/A pair would extract the chief complaint followed by the second Q/A pair that would extract the entity. E.g.: first Q/A pair: What is the chief complaint? A: The chief complaint is...  The second Q/A pair would be Q: When did it start? A: It started several hours prior to the arrival.

We observed that the extracted phrases (several hours prior to arrival) were unnecessarily verbose and consisted of extra tokens (in this case `to arrival'). On the advice of the physician, we developed a prompting template that would mimic a physician's thought process by urging the model to look for keywords that are commonly associated with the chief complaint and OPQRS entities. This helped to narrow down the scope for searching the entity in the long History of Present Illness (HPI).

The next step was to come up with an automatic way of extracting the generated outputs. Taking cue from~\citep{wang2023gpt}, we prompted the model to add `@' tokens before and after the final answer so that we could use regex parsing to extract the output to perform the large-scale evaluation.

After this, We involved the LLM in our prompt engineering process. We aimed to make the model understand our task by asking it the suitable way to prompt itself. We provided it with the task definition and asked it to refine our previously used reasoning steps. 
The model generated new reasoning steps that involved first extracting the `verb tense' and then extracting the chief complaint as opposed to extracting the chief complaint directly. It then generated other suitable rules to extract the other entities. We also used the LLMs to improve our entity definitions.



\subsection{Prompt Engineering}
Our proposed prompt is composed of six parts. The first part consists of a task definition that provides general guidelines to the LLM about the task, i.e. it is an NER task what is the entity we are looking for, and how to go about it. In our case, we extract the `Chief Complaint' first and then the entity. This is done so that we can exploit the dependency of the entity to the chief complaint.
The second part provides the definition of the `Chief Complaint' and the O/P/Q/R/S entity. 
The third part is where we try to mimic the physician's thought process. We ask the model to look for certain placeholder phrases to extract the `Chief Complaint' (Eg: `presents with', `complains of', `comes to/in', etc), and then the placeholder provides phrases corresponding to the entity of interest. (Eg: For `Onset' we will be looking at temporal markers like `several days/hours ago',  `today/yesterday', etc. ). The motivation behind this is that a physician when searching for the entities looks for these phrases first.
The fourth part is the reasoning steps that we provide that make the LLMs output their reasoning using the template we provided. Additionally, we noticed that the model outputs the correct reasoning during inference but falters at the end when producing the output, i.e. it is not able to consolidate its reasoning process to come to a conclusion. Hence we add a self-verification step to our reasoning steps that urges the model to go through its reasoning again before making its decisions. This method reduces hallucinations to a large extent.
The fifth part consists of a few-shot examples that have been carefully chosen to depict all sorts of combinations of chief complaint and the entity. The first example consists of an HPI note where the chief complaint and the entity both are present. The second HPI note example contains the chief complaint but no mention of the entity, while the third example does not contain both the chief complaint and the entity. 
Finally, the sixth part of our prompt consists of the instruction where we instruct the model to add `@' tokens before and after the final output for easy extraction.
All these parts were added to the prompt after extensive prompting and interactions with the LLMs. 


\subsection{Evaluation Metrics}
For traditional NER evaluation metrics, we adopt the evaluation protocol of SemEval 2013 task 9.1 ~\citep{segura2013semeval} to evaluate models for mention extraction and entity linking. Precision, Recall, and F1 scores are reported in ``Exact'' mode (exact boundary match) to measure the performance of entity extraction.

To allow for a more comprehensive assessment of the model's performance in cases where the generated text is not an exact match to the original, We extend the entity-based F1 score by proposing a new metric that measures the similarity between the generated phrase and the ground truth using BERTScore or PromptLLM.

The evaluation prompt was designed after performing extensive conversations between the human and the LLM. This was done to understand whether the LLMs can follow human-designed instructions. Mainly it was accomplished in 5 steps.
In step one, a simple definition of the task that can easily be understood by a human is provided to the language model. The human asks the LLM to respond with Yes/No if it understands the task. After ascertaining it has understood the task, for step two we provide a test instance that was designed keeping in mind the edge cases where the LLMs are more likely to generate errors. We observe the LLMs perform poorly on the example and are not able to reason out correctly. The performance was gauged by eyeballing the results and reasoning provided by the LLMs. In step three,  The LLM was then corrected by the human and the reasoning behind the correction was explained to it. Step 4 constituted drafting a question that would ask the LLMs to improve the task definition in a manner so that it does not commit similar mistakes again. The LLMs provide the refined definition which is in turn used to prompt the LLMs and to extract the entity in step five. 

\section{Experiments}

\subsection{Dataset}
The annotation process was carried out by the collaborating expert physician who is well-versed in understanding EHR data. An annotation guideline was created to lay down the rules. Some of the rules followed are:
To make sure that the ground truth is faithful to the source text we chose to keep any spelling/grammatical errors in the annotation.
If any entity does not exist the ground truth the annotation is an empty string.
We try to keep our annotation succinct and to the point by not including any unnecessary phrases. But we also make sure the phrases are complete in meaning.
If there are multiple mentions of an entity then we replicate the instance and create new data points per unique extracted phrase.
We took care to consider annotating OPQRS entities that are associated with the chief complaint only and not describing any other symptom. 

The annotation process resulted in a test set of 85 samples. The distribution of samples across all entities was as follows. The entity `Onset' had 63 instances followed by `Provocation' having 35 instances. `Palliation' had 20 instances and `Quality' had 53. `Region' had 49 followed by `Radiation' having 42 annotations. Finally `Severity' had the least number of instances i.e 16. Each dataset sample consists of the History of Present Illness (HPI) section of the EHRs along with the extracted phrase and its corresponding label (Onset, Provocation, etc). All data will be released under restrictions.

\begin{table}[t]
  \centering 
  \caption{Performance of the automatic evaluation pipeline with human evaluation}
  \begin{tabular}{lllllll}
  \toprule
    \textbf{Entity} & \textbf{Kappa-PromptLLM } & \textbf{Kappa-BERTScore} & \textbf{F1-Human} &  \textbf{F1-LLM}\\
    \midrule
     Onset & \textbf{0.836} & 0.646 & 0.948 & 0.944 \\
     Provocation & 0.788 & \textbf{0.793} & 0.877 & 0.836 \\
     Palliation & \textbf{0.869} & 0.832 & 0.943 & 0.941 \\
     Quality & \textbf{0.745}  & 0.695 & 0.704 & 0.673 \\
     Region & \textbf{0.880} & 0.690 & 0.754 & 0.736 \\
     Radiation & \textbf{0.923} & 0.897 & 0.903 & 0.889 \\
     Severity & \textbf{0.946} & 0.927 & 0.849 & 0.838 \\

    \bottomrule
  \end{tabular}
  \label{tab:main} 
\end{table}

\subsection{Experiment settings}
All our experiments were performed using the Llama-2-13B-chat model. We used one 80GB A100 for loading the model and carrying out inference. We used the Hugging Face Accelerate library for the deployment. With regards to the experiment settings, the temperature was set to a mid value of 1.0, the top\_p parameter was set to 0.95 and the top\_k parameter was set to 30. The combination of top\_p and top\_k was selected based on experimentation and observing the outputs. 
We want to make sure the model follows the reasoning steps but also is able to have certain flexibility in terms of generating tokens for the extracted entities. To limit the variety of tokens in terms of generating the reasoning steps we set the top\_k parameter to 30. To encourage variety in the extracted tokens in the reasoning steps we set a high top\_p value.
Depending on the stage of the pipeline, during the extraction phase, the maximum number of tokens was set to 2000. For evaluation, we set the maximum number of tokens to 500 as the reasoning was shorter in length as compared to the extraction phase.






\section{Results}


\begin{table}[t]
  \centering 
  \caption{F1 score of different prompting methods by evaluation pipeline (PromptLLM). The best/2nd-best scores in each column are in \textbf{bold}/\underline{underlined}.}
  \begin{tabular}{lllllll}
  \toprule
    \textbf{Entity} & \textbf{Prefix} & \textbf{Cloze} & \textbf{Anticipatory} &  \textbf{COT} & \textbf{Heuristic}   &  \textbf{Our method}\\
    \midrule
    Onset & 0.424 & 0.578 & \underline{0.618} & 0.488 & 0.303 & \textbf{0.944}\\
    Provocation & 0.18 & 0.247 & 0.16 & \underline{0.762} & 0.752 & \textbf{0.836} \\
    Palliation & 0.182 & 0.143 & 0.47& 0.695 & \underline{0.85} & \textbf{0.941} \\
    Quality & 0.235 & 0.400 & 0.351 & 0.163 &\underline{0.603} & \textbf{0.673} \\
    Region  & 0.464 & 0.302 & 0.351 & 0.346 & \underline{0.667} & \textbf{0.736} \\
    Radiation & 0.295 & 0.517 & 0.152 & 0.409 & \underline{0.68} & \textbf{0.889} \\
    Severity & 0.158 & 0.179 & 0.082 & 0.079 & \underline{0.478} & \textbf{0.838} \\
    \bottomrule
  \end{tabular}
  \label{tab:baseline} 
\end{table}

\begin{figure}[t]
  \centering 
  \includegraphics[width=6.5in]{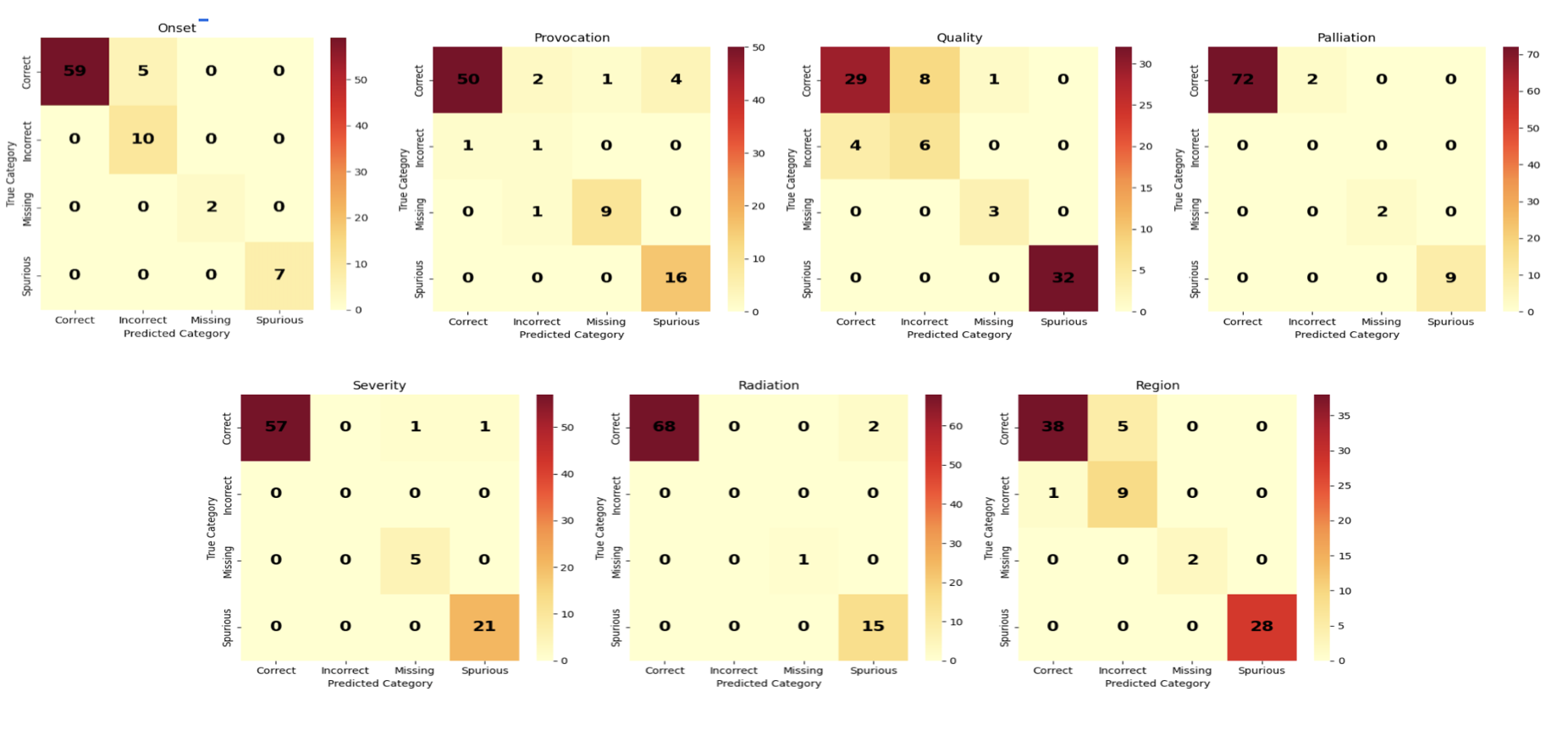} 
  \caption{Heatmap of OPQRS breakdown results.}
  \label{fig:heatmap} 
\end{figure}

Table \ref{tab:main} shows the performance of the proposed automatic evaluation pipeline along with human evaluation based on Cohen's kappa coefficient. The prompt-based method outperforms the BERTScore-based evaluation in most cases. Since PromptLLM evaluation can always achieve a Kappa score over 0.7, which indicates a very high agreement with human evaluation, we used PromptLLM evaluation instead of human evaluation for all the extraction evaluations in this work.

\begin{table}[t]
  \centering 
  \caption{We evaluate the efficacy of the proposed reasoning-based prompting template and the self-verification step for the entity `Onset'}
  \begin{tabular}{ll}
  \toprule
    \textbf{Prompt Method} & \textbf{F1}  \\
    \midrule
    Full prompt & 0.94  \\ 
    \quad- reasoning steps & 0.84  \\ 
    \quad\quad- self-verification & 0.769  \\ 
    \bottomrule
  \end{tabular}
  \label{tab:ablation} 
\end{table}

Table \ref{tab:baseline} presents a comparison between our prompting method and the techniques outlined in the study by \cite{info:doi/10.2196/55318}. This reference work conducts an extensive comparative analysis of various prompting techniques specifically tailored for clinical natural language processing tasks, including clinical sense disambiguation and biomedical evidence extraction. Given its relevance to our task, \cite{info:doi/10.2196/55318} paper serves as an ideal baseline for assessing the efficacy of our prompting method relative to established techniques in the field. For more information about each type of prompt refer \cite{info:doi/10.2196/55318}. Our study revealed some interesting observations. First, we observe that for most of the entities like `Palliation', `Quality', `Region', `Radiation', and `Severity' the heuristic prompts performed the best. Heuristic prompts contained the reasoning steps as rules to be used to extract the entity. This bolsters the claim that reasoning steps are an effective way to extract entities and model the complex relationships among them. Without reasoning steps, we observed the LLM is not able to follow the instructions well and ends up adding the ‘@’ tokens incorrectly. COT way of prompting has been considered one of the best ways of prompting for logical tasks \cite{NEURIPS2022_9d560961}, but it is not able to perform well for our use case. This can be attributed to the fact the reasoning provided by COT is too simple to truly capture the dependencies between the entities and the step-by-step reasoning it entails. Moreover, our task is niche and it is likely it has not been encountered by the language model during its pretraining phase, hence it is also not able to come up with the correct rationale behind the extraction of the entity. Hence for complex entity extraction, it is beneficial to provide the reasoning steps explicitly.


\begin{figure}[t]
  \centering 
  \includegraphics[width=5in]{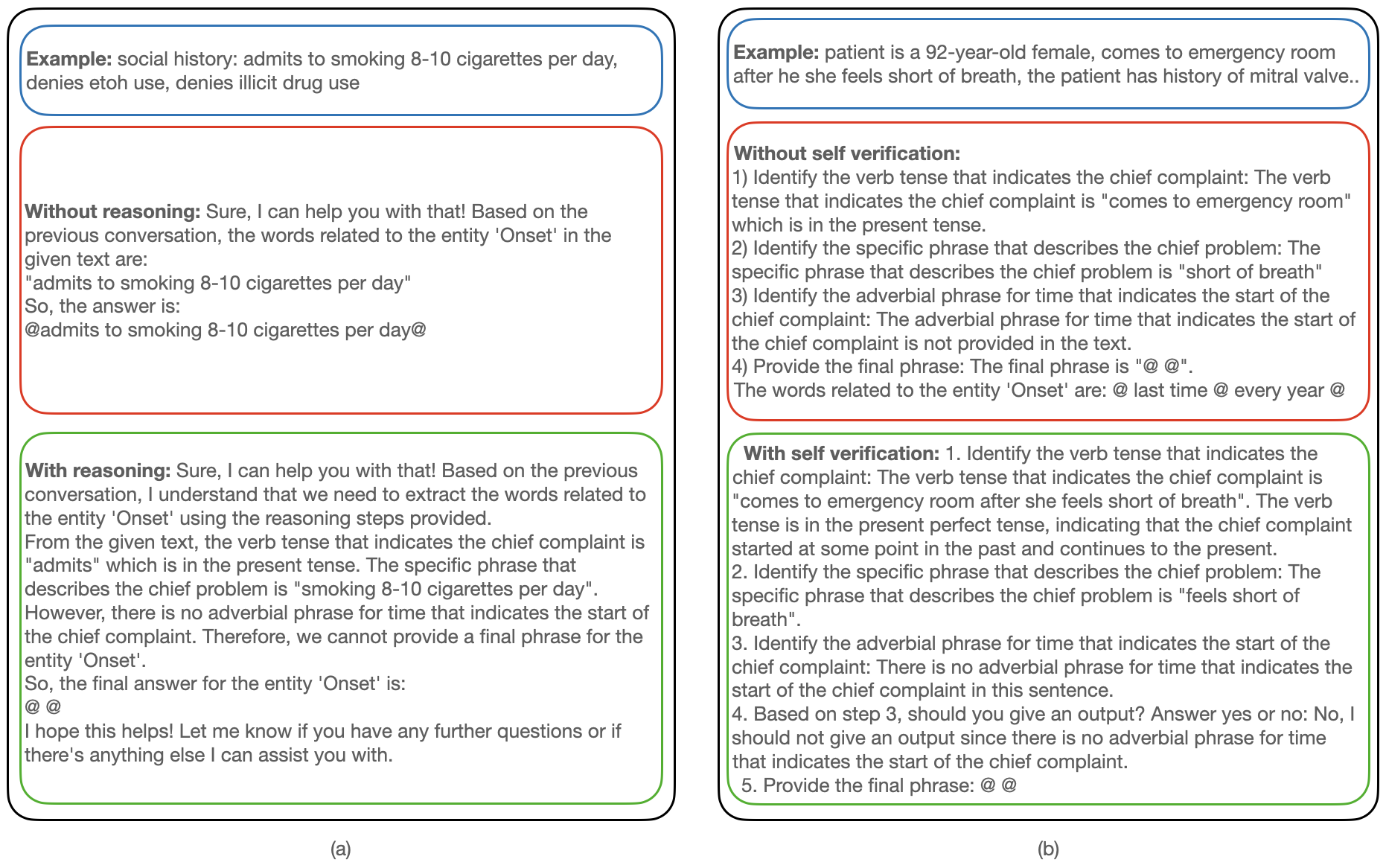} 
  \caption{Detailed prompts used in ablation study.}
  \label{fig:ablation} 
\end{figure}

\subsection{Ablation Study}
We are aiming to perform three ablation experiments to check the efficacy of our prompt.  The experiments are performed for the entity `Onset'. The first ablation experiment will gauge how adding reasoning steps influences the LLMs decision making process.  The reasoning steps were added to mimic the physician's thought process, i.e to first extract the chief complaint and then extract the entity `Onset'.  The prompt contains the definition of the entity and a corresponding few shot examples without the reasoning steps.
The second ablation experiment studies the effect of the self-verification step in our reasoning steps. The motivation behind adding the self-verification step was to mitigate hallucinations in the model's outputs. The prompt contains the definition of the entity and a corresponding few shot examples that contain the reasoning steps without the self-verification step. 
We compare these two settings to our method which contains the reasoning steps and the self-verification step.

If we compare the first two ablation experiments, The addition of reasoning steps is the difference. And that leads to a 7\% increment in scores. Figure \ref{fig:ablation} (a) is an example of how the prediction changes with and without the reasoning steps. We observe that the LLM is able to reason out better and comes to the correct answer with the reasoning steps. Another thing to note is that the F1 score jumped from 0.86 to 0.94 from ablation 2 to our method, i.e. an increment of 8\% more than before which highlights even if LLMs can reason out well they falter at the end due to hallucinations. Figure \ref{fig:ablation} (b) is an example of how the LLM output changes if we add the verification step.

\section{Discussion}
In this study, we introduced a novel approach for extracting OPQRST data from EHRs utilizing Large Language Models (LLMs) and reframing the task from traditional sequence labeling to text generation. Our research aimed to harness the sophisticated capabilities of LLMs to mimic the reasoning process of clinicians, thereby enhancing the interpretability. This transparency is crucial for clinical acceptance and trust, as it provides healthcare professionals with insights into the AI's decision-making process. On the other hand, We proposed and implemented modifications to traditional Named Entity Recognition (NER) metrics to better evaluate the performance of text generation models in a clinical context. By incorporating measures of semantic similarity, our approach provides a more comprehensive assessment of model output, recognizing the clinical intent of the text rather than just exact matches. This innovation helps in evaluating the true usability of AI-generated text in clinical practice.

While our results directly show the effectiveness of extracting OPQRST data, we believe this approach can also be applied to other healthcare settings where there is a lack of labeled training data, and transparency and decision-making processes are highly evaluated.




\section{Limitations}
This work is still limited in several ways. First, only in-context learning is used in this study, and we did not consider LLMs that are fine-tuned in the clinical corpus. Second, during the annotation process, we do not provide the reasoning steps to the physician. It would be interesting to explore the collaboration of Human-LLMs in this setting where the annotator improves their annotation by viewing the reasoning and the entity extracted by the LLMs while the LLMs inturn improve its suggestions garnering feedback from the humans.



\bibliography{main}

\newpage
\appendix

\section*{Baseline Prompts}
Here we provide the baseline prompts for our use case adapted from \cite{info:doi/10.2196/55318}

 \textbf{1. Prefix}:
In the EHR note, extract the words that describe the entity Onset, EHR note: ... \\

\textbf{2. Cloze}:
The words that describe the entity ‘Onset’ in the EHR note are .... EHR note: ... \\

\textbf{3. Anticipatory}:

Q1: What is Onset? A1: `Onset'  means the beginning or initiation of a symptom, sign, or medical condition.For ‘Onset’ we will be looking temporal markers. For example, `several days/hours ago',  `today/yesterday', etc. Q2: Extract the words that describe the entity ‘Onset’ in the EHR note?  EHR note: ... \\

\textbf{4. Chain of Thought (COT)}:

Extract the words that describe the entity ‘Onset’ in the EHR note: 
EHR note: This is a 73-year-old male comes in for evaluation of chest pain. he states the chest pain started about 7 or 7:30 this evening was off and on got progressively worse. he took some aspirin is morphine tablet and nitroglycerin at home with some relief of the chest pain but was still having its recall the ambulance committed for evaluation. he also had some mild shortness of breath. denied any nausea or vomiting. denied any diaphoresis. no lightheadedness or dizziness. denies any syncope or presyncope. no other complaints at this time. symptoms are moderate in severity. nothing is made this worse the nitroglycerin aspirated did seem to help somewhat

Answer: ‘7 or 7:30 this evening’ (tells us about the specific time frame when the chest pain started)

Question: Using the stored example extract the entity ‘Onset’ from the given EHR note
EHR note: ...\\

\textbf{5. Heuristic}:

Rules: 
Here are the rules to extract the entity ‘Onset’ from the EHR note.
1) Identify the verb tense that indicates the chief complaint.
2) Identify the specific phrase that describes the chief problem.
3) Identify the adverbial phrase for time that indicates the start of the chief complaint.
4) Provide the final phrase. 

Given the EHR note extract the entity ‘Onset’ using the rules mentioned above. EHR note: ...

\section*{Ablation Prompts}
Here we provide the prompts that we used for our ablation studies 1 and 2.

\subsection{Ablation prompt 1}

Definition: You are a chatbot who knows the task of `Named Entity recognition'. You are provided with some clinical data 
related to a patients visit to the emergency department of a hospital. The target task is to extract the the words
that belong to the entity `Onset'. `Onset'  means the beginning or initiation of a symptom, sign, or medical condition.
For ‘Onset’ we will be looking temporal markers. For example, `several days/hours ago',  `today/yesterday', etc. 

Now I will provide you with  few examples 

Example1: This is a 73-year-old male comes in for evaluation of chest pain. he states the chest pain started about 7 or 7:30 this evening was off and on got progressively worse. he took some aspirin is morphine tablet and nitroglycerin at home with some relief of the chest pain but was still having its recall the ambulance committed for evaluation. he also had some mild shortness of breath. denied any nausea or vomiting. denied any diaphoresis. no lightheadedness or dizziness. denies any syncope or presyncope. no other complaints at this time. symptoms are moderate in severity. nothing is made this worse the nitroglycerin aspirated did seem to help somewhat

Answer: @about 7 or 7:30 this evening@

Example2: 39-year-old white female presents with chest tightness. she relates that she was initially seen at a medics present was concern for blood clot so they sent her here. she relates she has a history of pulmonary embolus in her family. she has pain in her back at times. she denies any shortness of breath. she's had no syncope. she's had no recent illnesses.

Answer: @ @

Example3: Medications: per nursing records .

Answer: @ @

\subsection{Ablation prompt 2}

Definition: You are a chatbot who knows the task of `Named Entity recognition'. You are provided with some clinical data related to 
a patients visit to the emergency department (ED) of a hospital. The target task is to extract the the words that belong  to the entity `Onset'. For that we will first extract words related to the entity `Chief Complaint' and then based on  that we will extract the words related `Onset'. I will now give you the definition of both entities. `Chief complaint' is the main issue/problem that the patient presents to the ED with. `Onset'  means the beginning or initiation of a symptom, sign, or medical condition i.e the chief complaint (extracted before).

You have to make sure that you follow the definition I provide and not interpret the meaning of `Onset' in a general way.  So I will be providing with you some examples on how to extract the entities `Chief Complaint' and `{entity}. Basically we will be trying to mimic a physician's train of thought on how to extract these entities. 
We will be first looking for words that usually mark the starting of phrases related to `Chief Complaint'. For example, `presents with', `complains of', `comes to/in', etc and mostly the actual chief complaint follows  these words. For `Onset'  we will be looking temporal markers. For example, `several days/hours ago',  `today/yesterday', etc. \\

Putting this down into steps:

1) Identify the verb tense that indicates the chief complaint.
2) Identify the specific phrase that describes the chief problem.
3) Identify the adverbial phrase for time that indicates the start of the chief complaint.
4) Provide the final phrase. 

Now I will provide you with an example on how to extract the entity `Chief Complaint' and then the entity  `Onset' 

Example1: This is a 73-year-old male comes in for evaluation of chest pain. he states the chest pain started about 7 or 7:30 this evening was off and on got progressively worse. he took some aspirin is morphine tablet and nitroglycerin at home with some relief of the chest pain but was still having its recall the ambulance committed for evaluation. he also had some mild shortness of breath. denied any nausea or vomiting. denied any diaphoresis. no lightheadedness or dizziness. denies any syncope or presyncope. no other complaints at this time. symptoms are moderate in severity. nothing is made this worse the nitroglycerin aspirated did seem to help somewhat

Reasoning steps:
1) Identify the verb tense that indicates the chief complaint.- comes in for evaluation of chest pain 
2) Identify the specific phrase that describes the chief problem. - chest pain 
3) Identify the adverbial phrase for time that indicates the start of the chief complaint. -  chest pain started about 7 or 7:30 this evening  
4) Provide the final phrase.  - @about 7 or 7:30 this evening@\\

Now I will provide you with an example where the Chief Complaint is  present and but the entity `Onset' is not present. 

Example2: 39-year-old white female presents with chest tightness. she relates that she was initially seen at a medics present was concern for blood clot so they sent her here. she relates she has a history of pulmonary embolus in her family. she has pain in her back at times. she denies any shortness of breath. she's had no syncope. she's had no recent illnesses.

1) Identify the verb tense that indicates the chief complaint.- presents with chest tightness
2) Identify the specific phrase that describes the chief problem. - chest tightness
3) Identify the adverbial phrase for time that indicates the start of the chief complaint. - not provided
4) Provide the final phrase.  - @ @\\

Now I will provide you with an example where the Chief Complaint is not present and as a result the entity  `Onset' is not present as well. 

Example3: Medications: per nursing records .

1)Identify the verb phrase that indicates the chief complaint. -  not provided
2) Provide the final phrase. @ @

\end{document}